\documentclass[conference]{IEEEtran}
\IEEEoverridecommandlockouts
\usepackage[sort&compress]{natbib}
\usepackage{amsmath,amssymb,amsfonts}
\usepackage{algorithmic}
\usepackage{graphicx}
\usepackage{textcomp}
\usepackage{xcolor}
\usepackage{subcaption}
\usepackage{url}

\def\BibTeX{{\rm B\kern-.05em{\sc i\kern-.025em b}\kern-.08em
    T\kern-.1667em\lower.7ex\hbox{E}\kern-.125emX}}

\begin{document}

\title{Classification of radiology reports by\\
modality and anatomy: A comparative study}

\author{\IEEEauthorblockN{Marina Bendersky}
\IEEEauthorblockA{\textit{IBM Research} \\
San Jose, CA, USA \\
mbenders@us.ibm.com}
\and
\IEEEauthorblockN{Joy Wu}
\IEEEauthorblockA{\textit{IBM Research}\\
San Jose, CA, USA\\
joy.wu@ibm.com}
\and
\IEEEauthorblockN{Tanveer Syeda-Mahmood}
\IEEEauthorblockA{\textit{IBM Research}\\
San Jose, CA, USA\\
stf@us.ibm.com}
}

\maketitle

\begin{abstract}
Data labeling is currently a time-consuming task that often requires expert knowledge. In research settings, the availability of correctly labeled data is crucial to ensure that model predictions are accurate and useful. We propose relatively simple machine learning-based models that achieve high performance metrics in the binary and multiclass classification of radiology reports. We compare the performance of these algorithms to that of a data-driven approach based on NLP, and find that the logistic regression classifier outperforms all other models, in both the binary and multiclass classification tasks. We then choose the logistic regression binary classifier to predict chest X-ray (CXR)/ non-chest X-ray (non-CXR) labels in reports from \emph{different} datasets, unseen during any training phase of any of the models. Even in unseen report collections, the binary logistic regression classifier achieves average precision values of above 0.9. Based on the regression coefficient values, we also identify frequent tokens in CXR and non-CXR reports that are features with possibly high predictive power.

\end{abstract}


\section{Introduction}

Large data collections that can be comprised of text, images or even video, are becoming more easily available to researchers, clinicians and the public in general. It is quite often necessary, as a critical initial step, to mine input data before proceeding to further research or analysis.

In a research setting, careful and accurate data labeling can be a tedious and time-consuming task that often requires manual inputs and expert knowledge. Moreover, the same dataset might need to be relabeled multiple times, not only in cases where the same dataset is used for different research purposes but also in cases where the data is \emph{mislabeled}. Mislabeled data \cite{BrodleyFriedl1999} produces in itself at least 2 new problems; first, the mislabeled data needs to be identified and differentiated from correctly labeled data \cite{BrodleyFriedl1999,venkataramanEtAl2004}, and second, the mislabeled data should be corrected or removed from the dataset (if possible). Models trained with mislabeled data will most certainly yield low performance metrics, which raises yet again another question. Is the low performance of the model due to the model itself or related to the quality of the data and correctness of its labels?

In addition to the need to alleviate the general task of data labeling, we address the common problem of classifying radiology reports for which the type of procedure and/or the body anatomy imaged cannot be determined systematically, and, instead, require manual and detailed evaluation.

\begin{figure}
\centering
 \includegraphics[width=3.5in]{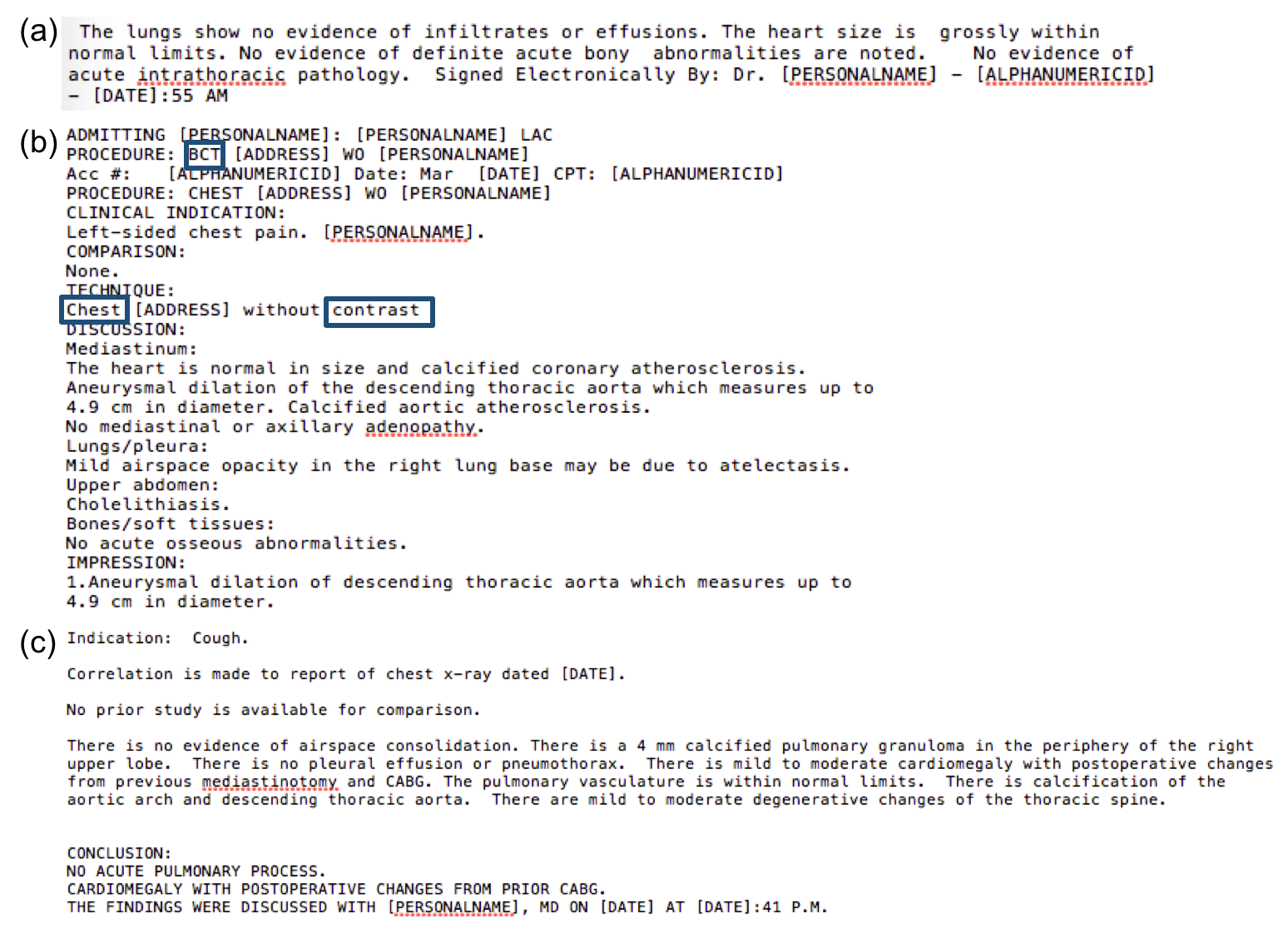}
 \caption{Examples of radiology reports in which the type of report cannot be identified systematically. From the manual analysis of the text, we infer that report (a) describes a CXR procedure and report (b) refers to a chest CT, while the classification label of report (c) is unclear.}
  \label{fig:reportsComparison}
\end{figure}

For example, the imaging modality of the exams described in the reports shown in Fig. \ref{fig:reportsComparison} can only be inferred by manual examination of the report text itself. The descriptions of lungs, heart and bones in Figs. \ref{fig:reportsComparison}(a) suggest the imaging modality is a CXR, though this is not explicitly included in the text and therefore the label could not have been inferred with a simple keyword matching approach. The structured text report shown in Fig. \ref{fig:reportsComparison}(b) suggests the exam is a \emph{chest CT}, due to the presence of the words ``CT", ``Chest", and ``contrast". Also in this case, it is not straightforward to obtain the ``chest CT" label with a keyword matching approach since the relevant tokens are not contiguous and are mentioned in different sections of the report. In particular, the relevant text ``chest without contrast" in the TECHNIQUE section is ambiguous, since it could imply a \emph{chest CT without contrast} or a \emph{chest MRI without contrast}. The example report shown in Fig. \ref{fig:reportsComparison}(c) is ambiguous in itself. On one side, it could correspond to a CXR report (a chest X-ray is mentioned as a reference study), though the content of the report itself describes a ``4 mm calcified pulmonary granuloma" which is most often and more clearly visualized in a chest CT.

The objective of this work is therefore to classify radiology reports by inferring \emph{jointly} the imaging modality of the procedure and the body anatomy being imaged. In the absence of a structured reports database, or when reports become available without the corresponding images, the automatic extraction of imaging modality and anatomy imaged is an essential first step that enables, for instance, a quick determination of follow-up procedures or treatment. The extracted information is also useful in the development of applications that improve the clinical workflow, such as summarization or medical information retrieval tools.

We propose a number of classification algorithms, based on machine learning and on NLP, that achieve high performance in both binary and multiclass classification tasks. The machine learning algorithms are relatively simple and implemented with open source libraries, such that researchers with any level of machine learning expertise can implement these ideas for their particular projects. Our models yield an average F1 score of around 0.9, so they could readily help save resources that would otherwise be spent on the expensive task of data labeling and classification. These models can also provide simple classification baselines to be used for comparison in the development of more complex approaches.

\section{Related work}

Many research studies focus on specific information extraction tasks that can improve clinical workflows or the performance of automatic tools already incorporated in those workflows \cite{infoExtractionReview2018}. For example, Yetisgen-Yildiz et al. \cite{yetisgenEtAlAMIA2011, yestisgenEtAlBMI2013} developed models based on NLP and on supervised learning to identify follow-up recommendations in radiology reports of different imaging modalities. Tiwari et al. \cite{tiwariEtAl2017} compared the performance of machine learning based models to classify findings according to their severity, in order to identify actionable findings indicative of the need for urgent clinician attention. 

Other researchers implemented ML and NLP methods to automate more general information extraction tasks. In particular, Chen et al. \cite{chenEtAlPEfindings2017} developed automatic methods to extract pulmonary embolism findings from thoracic CT reports, Gao et al. \cite{GaoEtAl2015} identified mammography findings by implementing a rule-based NLP approach, and Castro et al. \cite{castroEtAl2017} developed NLP and ML methods to automatically extract BI-RADS (Breast Imaging Reporting and Data System) categories.

A myriad of studies also focus on the analysis of reports for varied binary and multiclass classification tasks. For instance, Bijoy et al. \cite{bijoyEtAl2005} developed a string search strategy based on the boolean analysis of ankle X-ray reports to distinguish between different fracture cases, Massino et al. \cite{massinoEtAl2016} analyzed reports from the Audiological And Genetic Database to identify one or more ear regions of abnormality, and Shin et al. \cite{shinEtAl2017} developed a convolutional neural network (CNN) algorithm for the multi-task and multiclass classification of head CT reports. Classification of radiology reports can also enable retrospective studies. As an example of such an application, Zhou et al. \cite{zhouEtAl2014} developed NLP and machine learning techniques for the multiclass classification of \emph{sentences} found in the ``Impression" section of radiology reports of multiple imaging modalities and medical disciplines. Their study was not general, however, and instead confined to the analysis of reports that contained specific keywords related to sellar and suprasellar masses or to colloid cysts.
 
Researchers have conducted NLP and machine learning studies on reports of different imaging modalities or pertaining to different medical specialties. Each of these studies, however, is mostly concerned with the analysis of reports of \emph{only one} imaging modality and \emph{only one} body anatomy. Given the availability of \emph{large} collections of reports, comprising multiple imaging studies on multiple anatomies, we approach the broader problem of classifying the \emph{type} of reports, by identifying \emph{jointly} the imaging modality of the procedure and the body anatomy being imaged.

\section{Methods}
\label{sec:Methods}

We implement machine learning models to tackle the general problem of data classification (data labeling), which is often the immediate and essential task that researchers need to complete once large amounts of data become available to them. Depending on the specific research problem, researchers might be interested in identifying \emph{one} particular class or \emph{many} classes. For the cases in which we wish to distinguish only one class from the rest, it is appropriate to train binary classifiers that will distinguish between such class of interest and the rest of the classes as a whole. For cases in which there are multiple classes of interest, the analogous methodology consists in training multiclass classification models.

The data used for this comparative study comprises radiology text reports (no imaging data).  We approach the data classification problem by defining 2 tasks: the binary classification task, and the multiclass classification task. For the binary classification task, chest X-ray reports (CXR reports) are distinguished from non-chest X-ray reports (non-CXR reports). The methodologies (and results) are presented for the CXR/non-CXR classification in particular, but they can be generalized to other binary classification tasks. For the multiclass classification task, we define 21 classes of reports (such as Mammography, SpineMRI, ChestCT), and implement machine learning algorithms to efficiently distinguish between them.

We approach the binary classification task by implementing logistic regression, decision tree, and SVM (support vector machine) classifiers. We evaluate the performance of these classifiers both on training and testing data. We also introduce a NLP-based heuristic model and compare its performance to that achieved by ML-based classifiers. In the case of multiclass classification, we also implement logistic regression, SVM and decision tree classifiers. These experiments followed the ``one-vs-rest" paradigm \cite{RifkinKlautauOVR2004, multiclassAlgos2017}, which involves the fitting of $k=n\textsubscript{classes}$ binary classifiers.

\subsection{Machine Learning models}

The machine learning (ML) classifiers are implemented using the scikit-learn package in Python, trained on a dataset of 750 labeled reports and tested on a dataset of 250 labeled reports (as described in Sec. \ref{sec:dataPrep}). 

We consider two logistic regression classifiers, each of which is trained with either term-frequency (word-count) features or TFIDF  (term frequency-inverse document) features \cite{miningMassiveDatasets2014}. These models are regularized ($C=1$) and each of the classes are assigned equal weights. The decision tree classifier is trained with word-count features and each of the nodes in the tree is expanded until the leaves are ``pure", indicating that navigating a specific branch will lead to only one possible class (i.e., we do not impose a restriction on the maximum depth of the tree).

Lastly, we approach the binary classification task by training a support vector machine (SVM) classifier. A SVM is a learning model that constructs a hyperplane in the feature space such that the separation distance (or margin) between such hyperplane and the nearest data point of \emph{any} class is maximized. SVM classifiers are a type of kernel methods \cite{patternRecognitionBishop}, such that rather than learning fixed weight features, a SVM model learns a specific weight $w_i$ for the specific training example $\textbf{x}_i$. The prediction of an unlabeled test example is based on a similarity function $k$, or kernel, applied to the unlabeled example $y_{i}$ and the training example $x_i$. 

In our binary classification task, the predictions of the SVM classifier are based on a linear kernel given by
\begin{equation} 
k(x,y)=x^T y+c \,\,,
\label{eq:kernel}
\end{equation}
where $x$ is the training example, $y$ is the test example and $c$ is a constant. The first term in Eq.\ref{eq:kernel} represents the inner product $<x, y>$. The features used in our SVM classifier are word counts.

The models we trained for the multiclass classification task are also logistic regression, SVM (with a linear kernel), and decision tree. The models were trained with 701 labeled reports, as described in Sec. \ref{sec:dataPrep}, and the features are word-counts. All these models follow the paradigm ``one-vs-rest (ovr)", which consists of training $k$ binary classifiers where $k=n\textsubscript{classes}$. For each example report, the models yield its probability of belonging to each of the 21 classes and the predicted class is that with the highest probability. 

We also note the difference between ovr and one-vs-one (ovo) strategies \cite{HastieTibshirani1998, multiclassAlgos2017}. In the latter, binary classifiers are trained for each pair of classes, resulting in a more computationally expensive model since this involves the training of $\frac{k (k-1)}{2}$ classifiers. Given an example report, each binary classifier outputs a vote for either of 2 possible classes and the final prediction is derived from the majority of such votes. 

In our multiclass classification task, we tested the ovo approach and did not find a meaningful difference in performance with respect to ovr-based implementations. In terms of computational costs, however, the running time of the ovo approach is $~O(k^2)$, while for the ovr approach it is only $~O(k)$. The difference in running times increases rapidly with increasing number of classes ($k$). We therefore choose to perform all multiclass classification experiments following an ovr approach.

\subsection{NLP-based empirical model}
\label{sec:NLPmodel}

To obtain a baseline with which to compare ML-based predictions, we developed a NLP-based model that relies on the observed distribution of terms that are most frequently used in CXR reports. We observe a difference between the frequency of these terms in CXR reports and the respective frequency in non-CXR reports. In this section, we quantify such difference and derive a numerical threshold that can be used in our binary classification task.

Our training dataset of 750 labeled reports contains 81 CXR reports. The complete text in each of these reports was pre-processed as described in Sec. \ref{sec:dataPrep} and all the processed reports were joined to compose a ``CXR corpus". Within this CXR-corpus, we identify all the bigrams that appear at least 5 times and are composed of words that have at least 3 characters, none of which is numeric. We also obtain all the trigrams present in this corpus, though we do not impose a lower limit on their frequency since trigrams are repeated less frequently than bigrams. Each of the 3 words in each trigram has at least 2 characters, none of which are numeric. 

By detailed observation of the reports in our CXR-corpus, we compiled an arbitrary list of 63 ``CXR-terms" that were most frequently encountered in such reports. Some terms included in this list are ``chest", ``two", ``views", ``lung", ``clear", ``pleural", ``effusion", etc. We then proceeded to filter the bigram and trigram lists by selecting only bigrams and trigrams that contained at least one of the terms included in such list. We were then able to obtain bigrams and trigrams that are expected to be representative of text present in CXR-reports. Our analysis resulted in 99 ``CXR-bigrams" and 1,327 ``CXR-trigrams".

In order to train our NLP-based algorithm, we calculated the percentage of CXR-bigrams and CXR-trigrams in each of the 750 reports in our training dataset. This percentages simply represent the proportion of CXR-bi(tri)grams out of all bi(tri)grams present in the report, whether it was labeled as a CXR report or as a non-CXR report. For each type of report and each type of CXR N-gram we show in Tab. \ref{tab:cxrNgrams} the percentages of CXR N-grams found experimentally in our training dataset. 

As seen in Tab. \ref{tab:cxrNgrams}, in non-CXR reports the average proportion of CXR N-grams remains fairly constant and those proportions vary strongly between reports, given the high standard deviation. In CXR reports, the standard deviations are not only lower than the averages for both types of CXR N-grams, but we also observe that the average percent of CXR trigrams (0.423) is almost double than the average percent of CXR bigrams (0.222). 

In addition, we perform a two sample t-test to reject (or fail to reject) the null hypothesis that the mean percentage of trigrams in CXR-reports equals the respective value for non-CXR reports (i.e., $\mu\textsubscript{CXR, tri-grams} = \mu\textsubscript{non-CXR, tri-grams}$). The sample sizes of CXR and non-CXR reports are $n\textsubscript{CXR} = 81$ and $n\textsubscript{non-CXR}=669$, and the average percentages and sample standard deviations of trigrams are presented in the third column in Tab. \ref{tab:cxrNgrams}. The two-tailed test at the 5\% significance level yields a $p\textsubscript{value}\ll0.05$, such that we can reject the null hypothesis that the mean percents of trigrams are equal in CXR and non-CXR reports.

To derive a numerical threshold that can be used in the classification of unseen reports, we calculate (for each type of N-gram) the gap between the \emph{upper limit} in non-CXR reports and the \emph{lower limit} in CXR reports. Since this gap is wider in the case of trigrams, we calculate the mid-point of such gap and obtain a value of about 0.17. Based on this numerical limit, we then establish an empirical threshold for our binary classification task of 0.2.

Namely, for each unseen radiology report, we calculate the percent of CXR-trigrams (out of all trigrams present in the report). If the percent of CXR-trigrams is greater than 20\%, the unseen report will be classified as a CXR report; otherwise, it will be labeled as a non-CXR report.  

The presented NLP-based model is solely applied to the CXR/non-CXR binary classification and is not extended in this work to the multiclass classification problem. Such an extension is technically straightforward. It requires, however, the availability of a dataset with multiple types of radiology reports (preferably the same 21 classes defined in this work) and also the time-consuming definition of class-specific vocabularies and empirical thresholds. Those definitions could become a prohibitive task when $n\textsubscript{classes}=21$ and even more so as the number of classes increases.

\begin{table}
\centering
\begin{tabular}{ |c|c|c| } 
 \hline
 Report type & CXR-bigrams (\%) & CXR-trigrams (\%)  \\ 
 \hline
 chest X-ray & 0.222 $\pm$ 0.0953 & 0.423 $\pm$ 0.103 \\ 
 other & 0.0141 $\pm$ 0.0185 & 0.0142 $\pm$ 0.0203 \\ 
 \hline
\end{tabular}
\caption{Experimental distribution of CXR N-grams for each report type and each N-gram type, presented as intervals of (mean $\pm$ std. dev).}
\label{tab:cxrNgrams}
\end{table}

\subsection{Data pre-processing and construction of feature vectors}
\label{sec:dataPrep}

The radiology reports analyzed in this work are text files, and may sometimes contain different sub-sections such as ``History", ``Findings", or ``Impression". To avoid overfitting our models to the specific style adopted by particular radiologists or medical institutions, we analyze the complete text report. 

From a collection that includes $\sim$ 140,000 radiology reports, we randomly chose 1,000 of those reports and manually assigned a label to each of them. The reports in this collection were collected from 8 different imaging centers. To account for differences in report writing structures and styles, and in the frequency of specific exams (such as CXR, Chest CT, Mammography, Spine MRI) performed at each imaging center, we verified that the distribution of reports from each imaging institution in the original dataset was preserved in the manually labeled sample of 1000 reports.

For the binary classification task, we assigned ``CXR" and ``non-CXR" labels to each report in our sample of 1,000 reports. The non-CXR reports described studies performed using a different imaging technique and/or a different anatomy. We note this is an imbalanced dataset, with roughly $\sim$ 10\% of CXR reports. The binary-labeled reports were then randomly split into training and testing datasets, which contained, respectively, 750 and 250 reports. We also verified that the proportions of CXR reports in the training and testing datasets were $\sim$ 10\%, in agreement with the respective proportion in the complete labeled sample of 1,000 reports.

The pre-processing of the report text included the removal of punctuation, of non-alphanumeric characters, of de-identification tags, and of common headers and footers. The model features are either term-frequency (word count) features or TFIDF (term frequency - inverse document frequency) features. To obtain term frequency features, the processed text was converted to a sparse matrix of token counts and the resulting sparse matrix contained 7,826 word-count features. Using TFIDF features, each of the word counts was replaced by the scaled frequency of the term, and also resulted in a sparse matrix of 7,826 features.

\begin{figure}
\centering
  \includegraphics[width=3.6in]{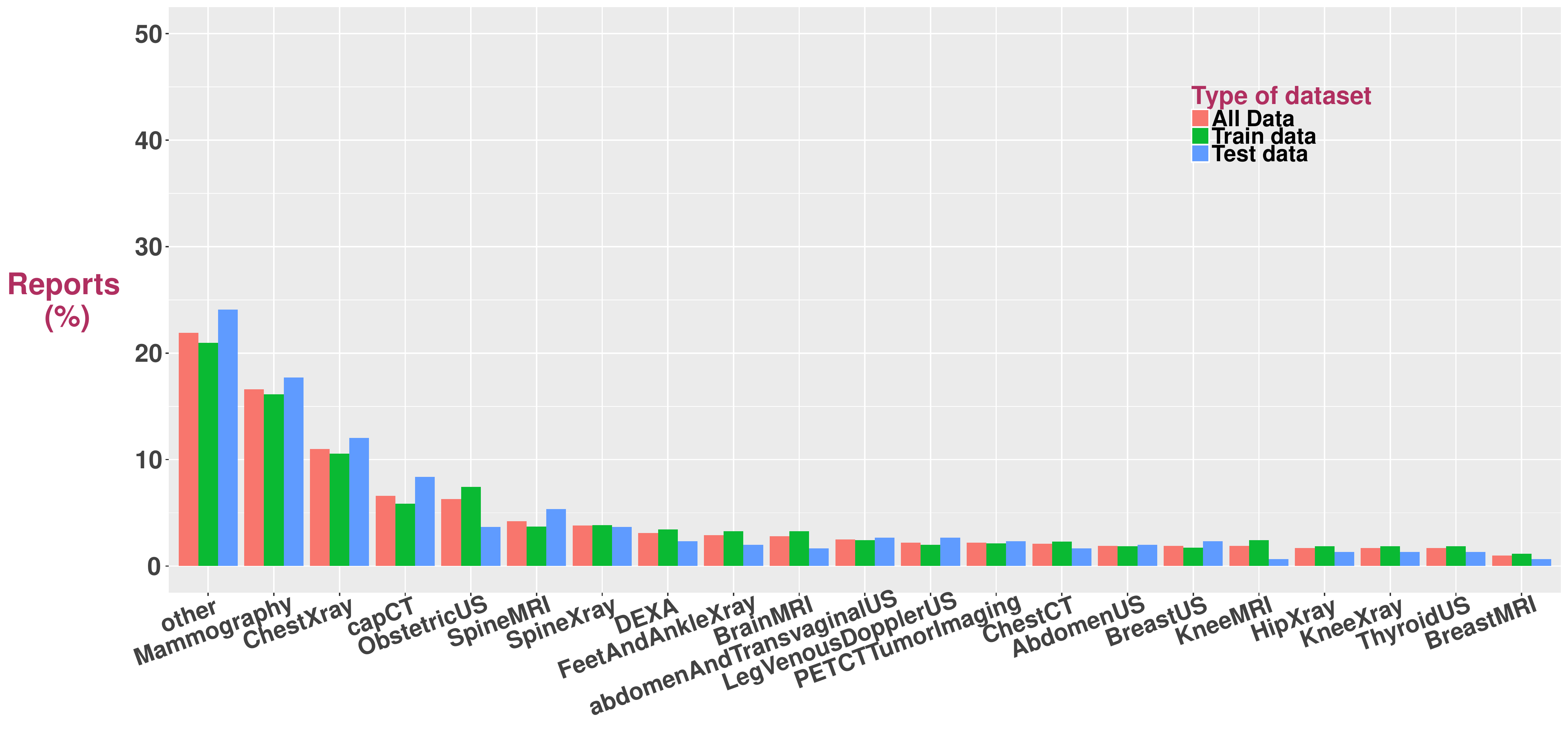}
  \caption{Distribution of the 21 report classes distinguished in the multiclass classification problem. Percents of reports of each class are presented for the complete dataset (1000 reports), the train data (701 reports) and the test data (299 reports). The abbreviation ``US" stands for ``ultrasound", ``DEXA" refers to Bone Densitometry Analysis, and ``cap CT" indicates ``chest, abdomen and pelvis CT" (distinguished from ``ChestCT").}
  \label{fig:mcDist21Labels}
\end{figure}

For the multiclass classification task, we define 21 classes of radiology reports. These classes were selected by re-labeling each of the non-CXR reports with a more informative label such as ``SpineXray", ``Mammography", or ``chestCT". We selected the 20 most frequent labels and the remaining reports, not belonging to any of these 20 classes were assigned to the category ``other". We split the complete set of 1,000 reports following a 70/30 random split, to obtain a training dataset of 701 files and a testing dataset of 299 files. We verified that this random split ensures an approximately equal representation of each class in each dataset, and the respective class distributions are shown in Fig. \ref{fig:mcDist21Labels}.

\section{Experimental results}
\label{sec:results}

After training all machine learning and NLP-based models, we evaluate the performance of each classifier in each of the binary and multiclass classification tasks of various datasets. We first report performance on the classification of the training and testing datasets, but we also expand the application of the classifiers to the analysis of other data collections, which were not seen during any training phase. Performance is evaluated by metrics such as precision, recall, F1 score and AUC (area under the curve). 

\subsection{Evaluation on the development dataset}

\subsubsection{Binary classification}

The performance of each of the binary classifiers was first evaluated in the training dataset (750 reports) with a 10-fold cross validation strategy. By iterating over the data 10 times, a cross-validation strategy allows for the calculation of confidence intervals for the precision and the recall, and thus provides more accurate estimates of the classifiers' performance on the test data. For each ML-based classifier described in Sec. \ref{sec:Methods}, we present the 95\% confidence intervals of the precision and the recall in Tab. \ref{tab:binaryTrainCV}. To evaluate the NLP-based approach, however, we do not perform cross-validation because this method is based on a single numerical threshold. In this case, confidence intervals are not relevant since we calculate the precision and the recall once, on the entire training dataset.

\begin{table}
\centering
\begin{tabular}{ |c|c|c| } 
 \hline
 Classifier & Precision & Recall  \\ 
 \hline
Logistic regression (word count features) & 0.96 $\pm$ 0.14     &   0.95  $\pm$  0.12\\ 
Logistic regression (TFIDF features) &  1.00 $\pm$  0.00   & 0.46   $\pm$ 0.44\\ 
Decision tree  &  0.90 $\pm$ 0.23  &   0.85  $\pm$  0.24\\ 
SVM (linear kernel) &  0.94  $\pm$   0.16  &     0.97  $\pm$   0.10   \\ 
NLP-based &   0.98     &       1.00      \\ 
 \hline
\end{tabular}
\caption{Precision and recall obtained in the binary classification of 750 training examples. For each algorithm, the results are averages of the performance in 10-fold cross validation and presented as 95\% confidence intervals (avg $\pm$ 1.96*std. dev). The NLP-based classifier was evaluated on the complete collection of 750 training examples (no CI is reported in this case).}
\label{tab:binaryTrainCV}
\end{table}

\begin{table}
\centering
\begin{tabular}{ |p{4cm}|p{0.95cm}|p{0.75cm}|p{0.6cm}|p{0.65cm}|} 
 \hline
Classifier & Precision & Recall  & F1  & AUC \\ 
 \hline
Logistic regression (word count features) & 1.00 &  1.00 & 1.00 & 1.00\\ 
Logistic regression (TFIDF features) &  1.00    &     0.67  &  0.8  & 0.83             \\ 
Decision tree  &   1.00   &     0.96     &    0.98   &      0.98    \\ 
SVM (linear kernel) &  0.96     &    1.00   &   0.98    & 0.998   \\ 
NLP-based &   0.81     &  0.96       &    0.88        &  N/A    \\ 
 \hline
\end{tabular}
\caption{Precision, recall, F1 and AUC scores for the binary classification of 250 test examples. The decision threshold in the NLP-based classifier is fixed at 0.20 so an AUC score is not defined in this case (reported as N/A).}
\label{tab:binaryTest}
\end{table}

Similarly to the results presented in Tab. \ref{tab:binaryTrainCV}, we report in Tab. \ref{tab:binaryTest} the performance metrics of those same classifiers on the test dataset of 250 reports. The metrics reported in Tab. \ref{tab:binaryTest} are precision, recall, F1 and AUC scores. A comparison of Tabs. \ref{tab:binaryTrainCV} and \ref{tab:binaryTest} shows that the cross validation values represent accurate estimates of the classifiers' performance in the testing dataset (which is the ultimate result of interest).

As reported in Tab. \ref{tab:binaryTest}, all classifiers yield high precision, with the lowest value of 0.81 obtained by implementing the NLP-based model. Recall values are also high (above 0.96), except for the low value of 0.67 obtained with the logistic regression classifier trained with TFIDF features. In agreement with previous work \cite{tiwariEtAl2017}, we obtain better classification performance with the use of word count features than with TFIDF features (compare the first 2 rows in Tab. \ref{tab:binaryTest}). While TFIDF features are useful in many contexts to decrease the weight of frequent words such as articles and prepositions, it is possible that the frequent words in a specific class of radiology reports are actually the differentiating factors between classes. Therefore, reducing the weight of these frequent domain-specific words might increase the similarity between reports that are \emph{actually} of different types. This effect could be related to the decrease in recall observed with the use of TFIDF features. Lastly, we note that the logistic regression classifier trained with word count features yields perfect performance metrics (all metrics are equal to 1).

\subsubsection{Multiclass classification}	

\begin{table*}
\centering
\begin{tabular}{ |c|| p{1.5cm} | p{1.5cm} | p{1.5cm} || p{1.5cm} | p{1.5cm} | p{1.5cm} | } 
 \hline
Classifier & Precision (micro-avg) & Recall (micro-avg)  & F1 score (micro-avg) & Precision (macro-avg)& Recall (macro-avg) &  F1 score (macro-avg)\\ 
 \hline
Logistic regression (word count features) &   0.9097  & 0.9097     & 0.9097   &   0.9202 & 0.893 & 0.896\\ 
SVM (word count features, linear kernel) &  0.906  & 0.906     & 0.906   &  0.899 & 0.896 & 0.887\\ 
Decision tree (word count features)  &  0.829  &  0.829 &  0.829 & 0.844 &   0.833 &  0.827  \\ 
 \hline
\end{tabular}
\caption{Precision, recall, and F1 scores for the multiclass classification (21 classes) of 299 test reports. For each classification algorithm, we report \emph{micro}-metrics, computed by considering all instances equally, and \emph{macro}-metrics, obtained by considering all classes equally.}
\label{tab:multiclassOnTest}
\end{table*}

\begin{figure}
\centering
  \includegraphics[width=3.45in]{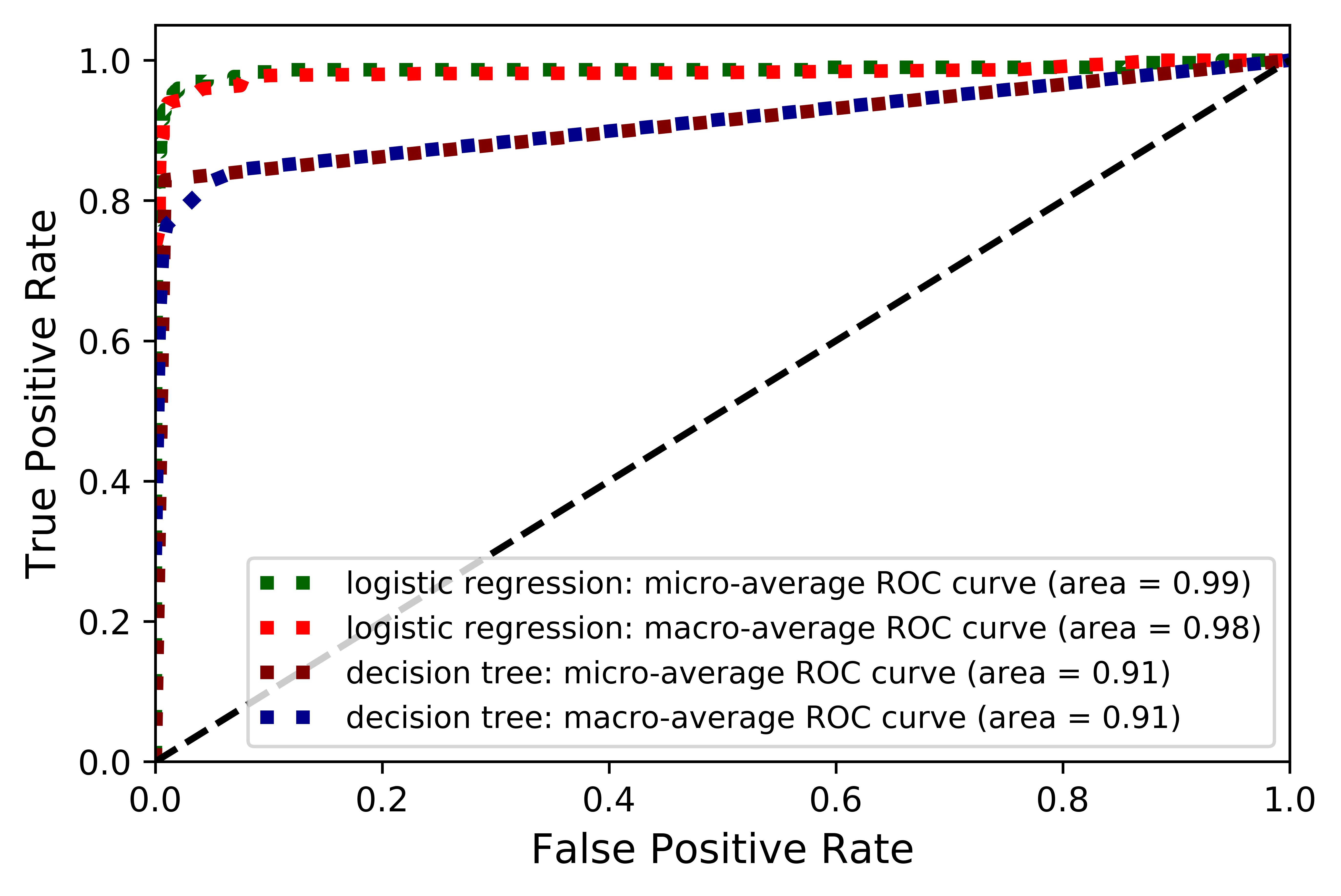}
  \caption{Micro- and macro-average ROC curves for the multiclass classification of reports with the logistic regression and decision tree classifiers.}
  \label{fig:mcROCComparison}
\end{figure}

The performance of multiclass classification algorithms on the test dataset of 299 reports is shown in Tab. \ref{tab:multiclassOnTest}. Metrics such as precision, recall and F1 scores are computed by either ``micro" or ``macro" averaging. A micro-average implies that the metrics were calculated globally for all instances, treating them equally and not taking into account their particular class. A macro-average, inversely, involves the calculation of metrics for each class separately followed by the unweighted mean of those metrics. The features in all of these multiclass classifiers are word counts and the SVM kernel is linear. 

The micro- and macro-performance metrics in Tab. \ref{tab:multiclassOnTest} show that the logistic regression and SVM models outperform the decision tree classifier. The linear classifiers (logistic regression, SVM) achieve a F1 score of $\sim$ 0.90, while the analogous value for the decision tree classifier is of $\sim$ 0.82. It is noted that such scores hardly vary with the specific type of averaging. A strict comparison of the micro- and macro-F1 scores achieved with logistic regression and SVM classifiers, indicates that the logistic regression classifier slightly outperforms the SVM classifier.

In Fig. \ref{fig:mcROCComparison}, we then compare the micro- and macro-average ROC curves obtained with the ``best" performing classifier (i.e., logistic regression) and the ``worst" performing classifier (i.e., decision tree). As expected from the metrics reported in Tab. \ref{tab:multiclassOnTest}, the AUC values for the logistic regression classifier are larger than those for the decision tree classifier, i.e. $\text{AUC (log. reg)}_\text{micro, macro}=0.99,0.98$ while $\text{AUC (dec. tree)}_\text{micro, macro}=0.91,0.91$.

\subsection{Evaluation on data from other sources}

In order to estimate the robustness of the \textbf{binary} classifiers more accurately, we choose a publicly available dataset that was not seen in any of the training phases of any of our classifiers. This dataset is part of the MIMIC database \cite{MIMICpaper2016}, which is broadly \emph{publicly} available, though access is restricted to users that complete a ``Data or Specimens Only Research" training course and that follow strict registration procedures \cite{MIMICwebsite2018}. Starting with the NOTEEVENTS table we selected only RADIOLOGY reports, and obtained a dataset of 522,279  reports. These reports also include a DESCRIPTION attribute, which is essentially the type of report, or, for our purposes, their classification label. Performance metrics for the binary classification of the MIMIC dataset are reported in Tab. \ref{tab:comparisonMIMIC}. We present precision, recall and F1 score values achieved with logistic regression, SVM, and NLP-based classifiers. As noted in Tab. \ref{tab:comparisonMIMIC}, the features of the ML-based models are word counts, and the SVM kernel is linear. 

\begin{table}
\centering
\begin{tabular}{ |p{4.64cm}|p{0.9cm}|l|l|} 
 \hline
 Classifier & Precision & Recall  & F1 score   \\ 
 \hline
Logistic regression (word count features) &  0.929   &  0.914 &  0.9214 \\ 
SVM (word count features, linear kernel) &      0.935      &  0.812    &     0.869   \\ 
NLP-based &   0.985     &      0.0249   &      0.0486         \\ 
 \hline
\end{tabular}
\caption{Comparison of the performance of logistic regression, linear SVM and NLP-based models applied to the binary classification of 522,279 labeled radiology reports that are part of the MIMIC database. We report in each case precision, recall and F1-score values.}
\label{tab:comparisonMIMIC}
\end{table}

As shown in Tab. \ref{tab:comparisonMIMIC}, the highest precision of $\sim$ 0.94 is achieved with the SVM classifier, while the logistic regression classifier yields the highest recall of $\sim$ 0.91. Overall, the best performance is achieved with the logistic regression classifier, which yields an F1 score = 0.92. 

The NLP-based model yields performance metrics that clearly exemplify the precision-recall trade-off \cite{patternRecognitionBishop}. In other words, the precision obtained in this case is extremely high and the recall is extremely low, to yield a remarkably low F1 score of roughly 0.05. As described in Sec. \ref{sec:NLPmodel}, this model is based on manually designed CXR-trigrams that were particularly fit to a specific dataset. We can be confident that, if these CXR-trigrams represent at least 20\% of the overall trigrams in the test report, the latter is indeed a CXR report. Inversely, our NLP-model does not consider all other (possibly infinite) CXR-trigrams that could be defined for other report collections, or, it is also possible that the threshold should be lower for such collections to reduce the number of false negatives. These observations could help explain the experimental results of very high precision and very low recall that we obtained in the classification of the MIMIC dataset.

As noted above, the results in Tab. \ref{tab:comparisonMIMIC} indicate that the best classification performance is obtained with the logistic regression classifier. Therefore, we select this classifier for the analysis of data collected from other sources, different from those of the data used to train our classifiers. For privacy reasons, we will refer to those collections as datasets `A', `B' and `C'. 

\begin{figure}
\centering
 \includegraphics[width=3.2in]{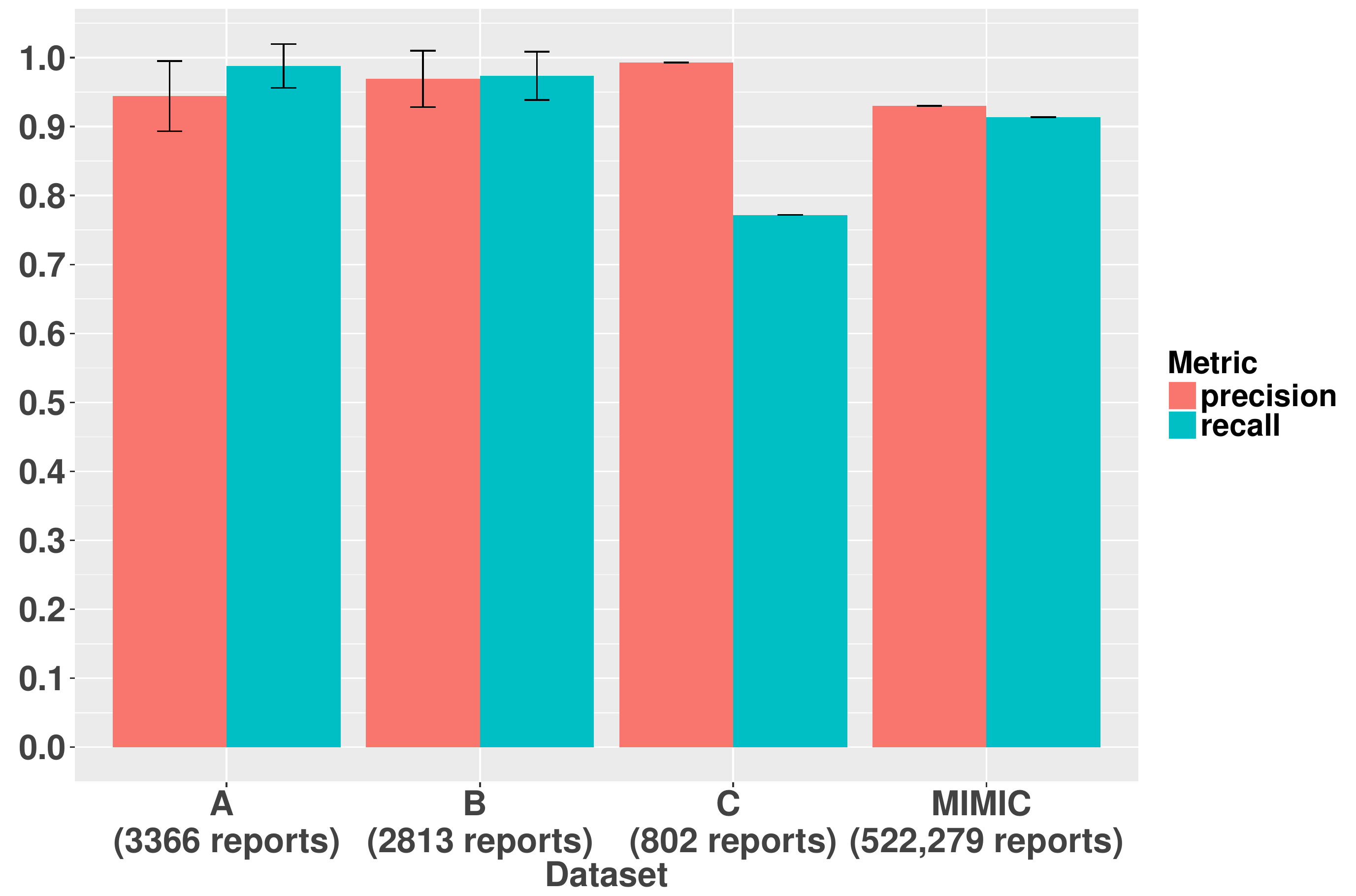}
  \caption{Performance of the logistic regression classifier (with word count features) on datasets from different sources. Precision and recall are presented as 95\% confidence intervals for datasets `A' and `B' because the metrics were calculated on random samples from the complete dataset. Datasets `C' and MIMIC are labeled so the metrics in those cases are computed for all of the reports.}
  \label{fig:logRegComparison}
\end{figure}

In Fig. \ref{fig:logRegComparison}, we show the performance metrics of the logistic regression classifier (with word count features) obtained in the classification of reports in the datasets `A', `B', and `C'. For a visual comparison, we also include the performance of the classification of the MIMIC dataset, which is also reported in the first row in Tab. \ref{tab:comparisonMIMIC}. 

Given the size of the \emph{unlabeled} collections `A' and `B', which comprised 3,366 and 2,813 reports respectively, we evaluate the classifier's performance by selecting 10 random samples of 100 reports each and labeling each of those reports manually. The precision and recall were calculated for each sample and the respective 95\% confidence intervals are shown in the bars labeled `A' and `B' in Fig. \ref{fig:logRegComparison}. Collection `C' and the MIMIC dataset are \emph{labeled} datasets, so random sampling is not required and the metrics were computed on the complete datasets. 

With the exception of the recall in the classification of dataset `C', we observe in Fig. \ref{fig:logRegComparison} that all other average metrics in datasets `A' and `B', and direct metrics in datasets `C' and MIMIC are above 0.9. The logistic regression classifier is therefore shown to be robust, since it yields high performance results, even on the classification of datasets from varying sources.

The classification of reports from varying sources described in this section refers only to the binary classification task. An extension of our work on multiclass classification to the analysis of reports from other sources is certainly relevant, though it could be limited by the availability of other data sources that present roughly the same 21 classes. We also infer that classification results might be affected if the distribution of classes vary significantly between the training and testing datasets \cite{datasetShiftBook2009}, being the latter from an entirely different source. The development of robust multiclass classifiers that can be efficiently applied in collections from varying sources remains as an interesting topic for future work.

\section{Discussion}

Experimental results presented in Sec. \ref{sec:results} indicate that radiology reports can be efficiently classified by implementing relatively simple ML-models. With the exception of the logistic regression model implemented with TFIDF features, we observe in Tabs. \ref{tab:binaryTest}-\ref{tab:comparisonMIMIC} that all F1 scores are above 0.82. 

The logistic regression classifier, implemented with word count features, outperforms all other classifiers in all of our experiments. It yields the highest F1 score in the binary classification of the test set (Tab. \ref{tab:binaryTest}), in the classification of the unseen MIMIC dataset (Tab. \ref{tab:comparisonMIMIC}) and in the multiclass classification of the test set (Tab. \ref{tab:multiclassOnTest} and Fig. \ref{fig:mcROCComparison}). The performance of the logistic regression classifier is followed by that of the SVM classifier (with word count features and a linear kernel), and lastly, by that of the decision tree classifier. Given the better performance of the linear classifiers (logistic regression, SVM), we infer that our data is actually linearly separable, which could explain the slightly lower performance of the decision tree classifier. 

The logistic regression \emph{binary} classifier is also shown to generalize quite well to the predictions of reports from \emph{different} collections, not seen during any development phase. This finding indicates that despite differences in writing styles and in report structures between radiologists and imaging institutions, CXR reports are characterized by a specific vocabulary that differs from the vocabulary commonly used to describe other types of exams, performed with other imaging modalities and on other anatomies.

The highest regression coefficients were assigned to words that strongly pushed the classification towards the positive class (CXR reports) while the presence of words with lowest coefficients biased the classification towards the negative class (non-CXR reports). 
The words with the highest and lowest regression coefficients in our binary classification task are shown in Tab. \ref{tab:tokensLogReg}. Such grouping of words is not surprising, since CXR reports are often referred to as \emph{chest two views}, and they often describe conditions such as \emph{pleural effusion}, \emph{pulmonary edema}, \emph{clear lungs}, or findings related to the \emph{cardiomediastinal silhouette}, or the presence/absence of \emph{pneumothorax}.
Inversely, reports of other imaging modalities, such as \emph{CT}, may include descriptions of numerical measurements and their units (such as \emph{cm}) and of the administration of \emph{contrast} medications that increase the resolution and diagnostic capability of the images produced.

\begin{table}
\centering
\begin{tabular}{ |p{1.4in}|p{1.4in}|} 
 \hline
 Words with highest regression coefficients & Words with lowest regression coefficients  \\ 
 \hline
chest, pleural, pulmonary, two, lungs, cardiopulmonary, clear, silhouette, pneumothorax & pain, contrast, seen, soft, was, cm, enlarged, there, ct \\
 \hline
\end{tabular}
\caption{Tokens detected in the binary classification CXR/non-CXR to which the classifier assigned the highest and lowest regression coefficients.}
\label{tab:tokensLogReg}
\end{table}

The empirical NLP-based model yields a relatively high F1 score=0.88 in the binary classification of reports in the test set (Tab. \ref{tab:binaryTest}). This finding is somewhat expected since the train and test datasets are part of the same report collection. However, when applied to the classification of reports in the MIMIC dataset \cite{MIMICpaper2016, MIMICwebsite2018}, the NLP-based model fails to generalize to unseen collections (i.e., it yields a remarkably low recall). Though this model is not an efficient classifier overall, it can still yield precise predictions with very low rates of false positives. The extreme results (very high precision, very low recall) therefore suggest that this model was overfit to the training dataset. 

The NLP-based model presented in this work is time consuming, subjective, dependent on human knowledge and experience, and most probably not scalable to more than a few classes. It is developed, however, to provide an example of a classification algorithm that is \emph{not} based on ML, and to provide a baseline for comparison with the performance of ML-based algorithms.

\section{Conclusions}

We developed relatively simple ML-based approaches to facilitate the classification of radiology reports into 2 or more classes. Among these ML-based models, we find the logistic regression classifier outperforms all other models in both classification tasks (binary and multiclass), achieving an average F1 score $>$ 0.9. It is also robust, since it yields high performance metrics in the classification of reports from 4 sources different from the development dataset. As a baseline for comparison with the ML-based models, we also develop an empirical NLP-based model that does not generalize well to unseen collections, since it yields very high precision and a remarkably low recall. 

Our work suggests that the classification of radiology reports to identify modality and anatomy can be quickly approached with simple ML-based models that do not require complex feature engineering nor the fitting of classifier parameters or decision thresholds. These models yield high performance metrics and can also serve as classification baselines for the development of more complex models.

\bibliographystyle{unsrt}
\bibliography{myBibliography}

\end{document}